\pdfoutput=1
\documentclass{article}

\usepackage[preprint,nonatbib]{neurips_2022}

\usepackage[utf8]{inputenc} 
\usepackage[T1]{fontenc}    
\usepackage{hyperref}       
\usepackage{url}            
\usepackage{booktabs}       
\usepackage{amsfonts}       
\usepackage{nicefrac}       
\usepackage{microtype}      
\usepackage{algorithm}
\usepackage{algpseudocode}
\usepackage{multirow}
\usepackage{amsmath}
\usepackage{graphicx}
\usepackage{float}
\usepackage{caption}
\usepackage{subcaption}
\usepackage[normalem]{ulem}
\useunder{\uline}{\ul}{}
\usepackage[table,xcdraw]{xcolor}
\usepackage{scrextend}
\DeclareMathOperator*{\argmax}{arg\,max}
\DeclareMathOperator*{\argmin}{arg\,min}

\newcommand{\cX}{{\mathcal X}}
\newcommand{\cN}{{\mathcal N}}
\newcommand{\cF}{{\mathcal F}}

\newcommand{\rR}{{\mathbb R}}
\newcommand{\rP}{{\mathbb P}}
\newcommand{\rE}{{\mathbb E}}
\newcommand{\ind}[1]{\ 1\hspace{-2.3mm}{1}{\left\{#1\right\}}}
\newcommand{\hDelta}{{\widehat \Delta}}
\newcommand{\htheta}{{\widehat \theta}}
\newcommand{\dist}{{\mathrm{dist_\star}}}

\newcommand{\cE}{{\mathcal E}}
\newcommand{\cT}{{\mathcal T}}

\newtheorem{theorem}{Theorem}
\newtheorem{lemma}{Lemma}

\algnewcommand\algorithmicforeach{\textbf{for each}}
\algdef{S}[FOR]{ForEach}[1]{\algorithmicforeach\ #1\ \algorithmicdo}

\title{A Contextual Bandit Approach for Learning to Plan in Environments with Probabilistic Goal Configurations}

\author{Sohan Rudra$^*$ \\ Google Research \And Saksham Goel$^*$ \\ Google Search \And Anirban Santara$^*$ \\ Google Research \And Claudio Gentile$^*$ \\ Google Research \And Laurent Perron \\ Google Research \And Fei Xia \\ Robotics@Google \And Vikas Sindhwani \\ Robotics@Google \And Carolina Parada \\ Robotics@Google \And Gaurav Aggarwal \\ Google Research}

\begin{document}

\maketitle

\begin{abstract}
Object-goal navigation (Object-nav) entails searching, recognizing and navigating to a target object. Object-nav has been extensively studied by the Embodied-AI community, but most solutions are often restricted to considering static objects (e.g., television, fridge, etc.). We propose a modular framework for object-nav that is able to efficiently search indoor environments for not just static objects but also movable objects (e.g. fruits, glasses, phones, etc.) that frequently change their positions due to human intervention. Our contextual-bandit agent efficiently explores the environment by showing optimism in the face of uncertainty and learns a model of the likelihood of spotting different objects from each navigable location. The likelihoods are used as rewards in a weighted minimum latency solver to deduce a trajectory for the robot. We evaluate our algorithms in two simulated environments and a real-world setting, to demonstrate high sample efficiency and reliability.
\end{abstract}

\section{Introduction}
Personal robotics is an important domain of Embodied AI \cite{pfeifer2004embodied} that aims at enhancing human productivity by developing physical assistants for everyday tasks. In this paper, we study the task of Object-Goal Navigation (object-nav) \cite{chaplot2020object,ye2020seeing}, which is a core component of many personal robotics applications like Mobile Manipulation \cite{ahn2022can}, Embodied Question Answering \cite{das2018embodied}, and Vision-and-Language Navigation \cite{anderson2018vision}\footnote{\label{website}Please visit our website: \url{https://sites.google.com/view/find-my-glasses/home} for a video presentation of the paper and real world demonstrations.}. Object-nav is defined as the task of searching (and optionally, retrieving) a given object within a designated space, e.g., indoor spaces like homes or offices, which are typically {\em known} environments. The target object is often specified in natural language, e.g., ``a blue soccer ball with stripes". This work is motivated by an everyday \emph{object-nav} task: searching for our keys or cellphone at home. Analogous to how humans search, we propose an algorithm that learns to look for objects that change their locations due to human intervention (e.g. cellphones, glasses and keys) in the most likely places first.
Object-nav algorithms work by learning semantic relationships between the target object and the topology of the environment. They can be classified into two categories: map-based and map-free \cite{bonin2008visual}. Map-free algorithms \cite{talukder2003real,santos1995visual,mousavian2019visual,chen2018learning, ramakrishnan2022poni, objectnavgoog, georgakis2022learning} do not require a map of the environment and can decide which way to go based directly on the current observations and past memories without having to maintain a global representation of the environment. This requires them to solve the problem of localization and mapping in conjunction with the object-nav problem, making their sample complexity very high. Another prominent challenge faced by end-to-end learning-based algorithms in robotics is bridging the sim-to-real gap \cite{hofer2021sim2real}. Learning agents are usually trained in a simulator (sim) like Matterport \cite{Matterport3D}, AI2Thor \cite{Kolve2017AI2THORAI}, Gibson \cite{xiazamirhe2018gibsonenv} and Habitat \cite{habitat19iccv} before deploying in the real world. However, due to limited fidelity of a simulator, observations of the same event might be different in the simulator and in reality (domain mismatch). 
Map-based algorithms \cite{meng1993neuro,meng1993mobile,pan1995fuzzy,kim1999symbolic,borenstein1991vector,borenstein1989real} assume that a map of the environment is available at the time of path-planning. The map could be an occupancy map showing the probability of obstacles being at each location. It could also be a topological map \cite{desouza2002vision} that is comprised of a graph where nodes represent characteristic places and edges contain reachability information (distances, times, etc.) between pairs of nodes. A few of these algorithms construct a map of the current region before planning in that region \cite{chaplot2020learning, chaplot2020neural, scenememorytransformer, chaplot2020object}. Map-based methods are typically modular, hence, sample-efficient \cite{shalev2016sample} and easier to deploy in the real world. 
However, the validity of the solution devised by these algorithms is a strong function of the accuracy of the map and localization of the robot. Thanks to advancements in Simultaneous Localization and Mapping (SLAM) algorithms \cite{dissanayake2011review}, constructing an occupancy map of reasonable accuracy has become relatively easy in modern robotic systems.

In this paper, we aim to achieve robust Object-Goal Navigation in indoor human-centric environments. Our algorithms are modular and map-based. They assume access to a binary 2D occupancy map of the environment with navigable and non-navigable parts marked out. 
In our day to day life, quite frequently, we have to search portable objects that we happened to lose track of. We will refer to such objects as ``movable objects". Let us consider the example of \emph{cellphone}. We begin our search by trying to recall a list of places that we are used to finding our \emph{cellphone} at. The algorithm presented in this paper takes a similar approach where we aim to model the likelihood of spotting a given object from different places in the environment.  For example, a \emph{cellphone} is more likely to be found on the study-table, work-desk and bedside-table, rather than in the kitchen floor or on top of the refrigerator. These likelihoods are learned online as the agent explores a realistic environment. For stationary objects, we just assign probability values of $0$ or $1$. These likelihoods are then used for planning a path that minimizes the average distance travelled by the agent to reach the target object.

\begin{figure}
    \centering
    \includegraphics[width=\textwidth]{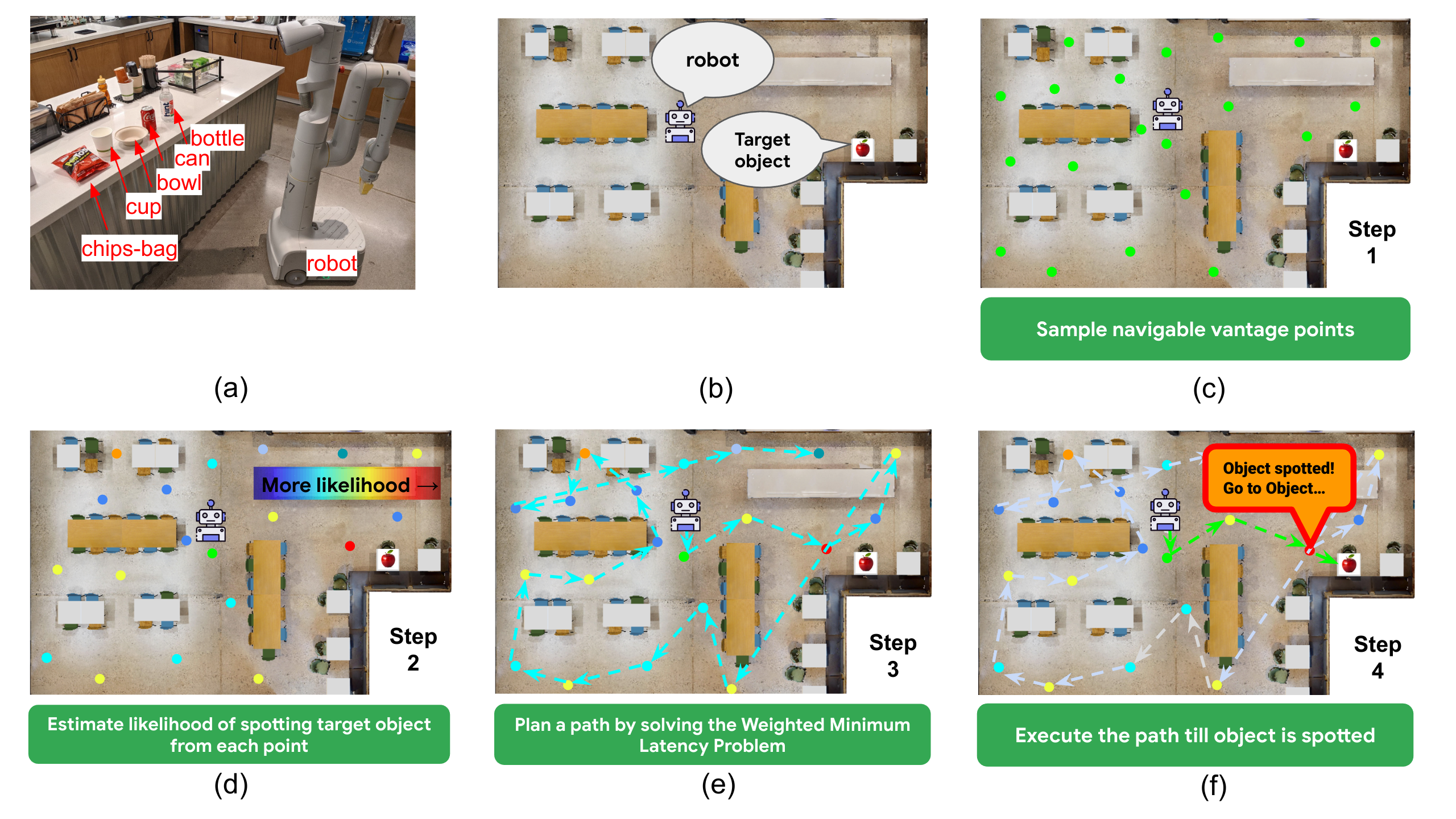}
    \caption{Overview of our approach. (a) Picture of our robot (from Everyday Robots) and the target objects studied in our experiment. (b) The robot is randomly initialized in the environment with the task of finding a given target object. (c) A set of reachable vantage points (green dots) are sampled across the entire environment using the current $2D$ occupancy map. (d) The Contextual-Bandit agent estimates the likelihood of spotting the target object from each vantage point. (e) A Weighted Minimum Latency Problem (WMLP) solver is used to generate an ordering of the vantage points taking into account their likelihood scores, the initial position of the robot and the geometry of the room. (f) The robot visits the vantage points in the planned order while inspecting its surroundings. As soon as it spots the object, it heads directly to it.}
    \label{fig:overview}
\end{figure}

Figure \ref{fig:overview} provides an overview of our approach. (1) The robot is randomly initialized in the environment with the task of finding a given target object. (2) A set of reachable vantage points are sampled across the entire environment using the current $2D$ occupancy map via farthest point sub-sampling \cite{eldar1997farthest,qi2017pointnet++}. (3) A Contextual Bandit Agent~\cite{li2010contextual} estimates the likelihood of spotting the target object from each vantage point. (4) A Weighted Minimum Latency Problem (WMLP) solver \cite{wei2018new} is used to generate an ordering of the vantage points taking into account their likelihood scores, the initial position of the robot and the geometry of the room. Finally, (5) The robot visits the vantage points in the planned order while inspecting its surroundings. As soon as it spots the object, it heads directly to it. 
The modular design of our approach significantly reduces the sample complexity and allows us to switch out object detectors, motion planners and point samplers for domain-specific models when an agent is transferred between sim and real -- reducing the sim-to-real gap.
The paper is organized as follows. In section \ref{s:theory}, we introduce the notation followed in the paper. We also provide a formal definition of our problem and a theoretical algorithm to address it. In section \ref{s:proposed_methodology}, we present a practical implementation of the theoretical algorithm. In section \ref{s:experiments}, we present an empirical evaluation of the proposed methods in two simulated and one real environments. In section \ref{s:limitations}, we conclude the paper with a discussion on the limitations of the proposed approach and directions of future work.

\section{Model and Algorithm}\label{s:theory}
%
We start by explaining the mathematical model underpinning our investigation.

\noindent{\bf Model.}
We formalize our problem as follows. In the 2D occupancy map of the environment, let us denote all the points by a set $\cX$. Set $\cX$ is partitioned into the set of {\em feasible} points $\cF$ (the points the robot can freely navigate) and the set of {\em non-feasible} points $\cN$ (the points occupied by obstacles like furnitures), so that $\cX = \cF \cup \cN$, and $\cF \cap \cN = \emptyset$. The three sets $\cX$, $\cF$, and $\cN$ are available to us before planning. 
%
Each $x \in \cF$ comes with a {\em visibility set} $V(x) \subseteq \cX$, that is, a set of points that the robot can inspect while standing\footnote
{
We are assuming here that the robot can sample from $x$ any pose via 360 degree rotation.
} 
at $x$. For concreteness, visibility is defined in terms of Euclidean distance as\ \  
\(
V(x) = \{x' \in \cX\,:\, ||x-x'|| \leq r_\mathrm{vis} \},
\)
\  
for some visibility range $r_\mathrm{vis} > 0$, e.g., $r_\mathrm{vis} = 2.5$ meters (the effective range of the object detectors on the system).

We have $n$ movable objects of interest (glasses, keys, etc.). We use $[n]$ to denote the set $\{1, 2, \dots, n\}$.
For each pair $(i,x)\in [n]\times\cF$, denote by $p_{i}(x)$ the probability that the robot spots object $i$ 
while standing at position $x$. These probabilities are in turn defined by $n$ (unknown) probability distributions $\{\rP_i$, $i \in [n]\}$, with support $\cX$, that determine where objects are located,
so that $p_i(x) = \rP_{y\sim \rP_i}(y\in V(x))$. 
Notice that an object can, in principle, be anywhere in the scene.
%
The probabilities $p_i(x)$ are unknown to the planner, and have to be learned through interactions with the environment. 
We model them as\ \ 
\(
p_{i}(x) = f(\phi(i,x);\theta),
\)
where $\phi\,:\,[n]\times\rR^d \rightarrow \rR^D$ is a mapping that featurizes the pair $(i,x)$ into a $D$-dimensional real vector, for some feature dimension $D \geq d$,
$f\,:\,\rR^D\times \rR^m \rightarrow [0,1]$ is a known function, and $\theta \in \rR^m$ is an unknown vector of parameters, for some parameter dimension $m$. For instance, 
$f(\phi;\theta) = \sigma(\theta^\top\phi)$, where $\sigma$ is the sigmoidal function $\sigma(z) = \frac{e^{zs}}{1+e^{zs}}$ with slope $s$ at the origin, and $m = D$. Our theoretical analysis uses the above generalized linear model,
while in our experimental evaluation, we compare both linear and neural models for $f(\cdot; \theta)$. 


\noindent{\bf Planning and Regret.}
In each navigation episode $t = 1, 2, \ldots$, there will be only one
object $i_t \in [n]$ in the scene\footnote
{
This is not a strict requirement, our analysis can be seamlessly extended to the case where multiple instances of the same object are simultaneously present on the scene.
}, 
and the identity of this object is {\em known} to the robot. The environment generates the position $y_{t}$ of object $i_t$ by drawing $y_{t}$ from $\rP_{i_t}$.
Let $x_{0,t} \in \cF$ be the starting position of the robot in episode $t$. The algorithm begins by sampling a total of $k$ vantage points $x_{1,t},\ldots, x_{k,t} \in \cF$. The planner has to generate a path $J_t = \langle x_{\pi_t(1),t},\ldots, x_{\pi_t(k),t} \rangle$ across them, where $\pi_t(\cdot)$ is a permutation of the indices $\{0\}\cup[k]$, with $\pi_t(0) = 0$. The robot traverses the path in the order dictated by $J_t$, and stops as soon as the object is spotted, as allowed by the visibility structure $V(x_{\ell,t})$, $\ell \in \{0\}\cup[k]$.
It is also reasonable to admit that the robot may incur some detection failures during an episode, an event we denote by $\cE_t$, to which we shall assign an independent probability $\rP(\cE_t)$ to occur.
We henceforth drop the episode subscript $t$ for notational convenience.

The {\em path length} loss $L(y,J)$ of $J$ is the actual distance traversed by the robot over the points $x_{0}, x_{1},\ldots, x_{k}$ {\em before} spotting object $i$ in position $y$. On the other hand, if the object is not found, it is reasonable to stipulate that the loss incurred will be a large number $L_M > \sum_{\ell = 1}^k \sum_{j=1}^\ell ,\dist(x_{\pi(j-1)}, x_{\pi(j)})$, bigger than the total path length of any length-$k$ path $\langle x_{\pi(1)},\ldots, x_{\pi(k)} \rangle$. Overall
\begin{align}
&L(y,J)= 
(1-\ind{\cE})\label{e:pathlength}\\
&\times\Biggl(\sum_{\ell = 1}^k \underbrace{\ind{y \in V(x_{\pi(\ell)}) \setminus (V(x_{\pi(0)}) \cup \ldots \cup V(x_{\pi(\ell-1)})  )}}_{\mbox{\begin{tiny}$V(x_{\pi(\ell)})$ is the first ball in the traversal order where object is located\end{tiny} }}\times\sum_{j=1}^\ell\,\dist(x_{\pi(j-1)}, x_{\pi(j)})\Biggl) + \ind{\cE}L_M~,\notag
\end{align}
where $\ind{\cdot}$ is the indicator function of the predicate at argument, and $\dist(x_1,x_2)$ denotes the $A^\star$ distance between the two vantage points $x_1$ and $x_2$ on the scene, i.e., the collision-free shortest path length between $x_1$ and $x_2$ for our robot (notice that $\dist(x_1,x_2) \geq ||x_1-x_2||$). 
Observe that, in the absence of a failure, we are assuming the object will eventually be found during each episode. Hence, a failure will be ascertained only at the end of an episode.
We approximate the above by disregarding the overlap among the visibility balls $V(x_{\pi(\ell)})$ (hence somehow assuming these balls do not influence one another), and then take expectation over $y \sim \rP_i$ and the independent Bernoulli variables $\ind{\cE}$, with expectation $\rP(\cE)$. This yields the (approximate) average path length
\vspace{-0.05in}
\begin{align}
&\rE_i [L(y,J)]\label{e:target} 
= \rP(\cE)\,L_M + (1-\rP(\cE))\Biggl(\sum_{\ell = 1}^k p_{i}(x_{\pi(\ell)})\sum_{j=1}^\ell\,\dist(x_{\pi(j-1)}, x_{\pi(j)})\Biggl)~,
\end{align}
where $\rE_i[\cdot]$ is a short-hand for $\rE_{y \sim \rP_i,\cE}[\cdot] $.
We are now in a position to define our benchmark performance measure against which regret will be defined.  The {\em benchmark planner} knows the distributions $\{\rP_i$, $i \in [n]\}$ and the probability $\rP(\cE)$, and computes a path
$J^\star = \langle x^*_1,\ldots,x^\star_k\rangle$ whose elements are taken from $\cF$, such that $J^\star$ minimizes $\rE_i[L(y,J)]$ over all length-$k$ paths (and permutations thereof) that can be constructed out of points from $\cF$. Given a sequence of episodes $t=1,\ldots, T$ with corresponding objects $i_1,\ldots, i_T$ (and starting positions $x_{0,1},\ldots, x_{0,T}$), we define the cumulative {\em regret} $R_T(J_1,\ldots,J_T)$ of a planner that generates paths $J_1,\ldots, J_T$ as
\[
R_T(J_1,\ldots,J_T) = \sum_{t=1}^T \rE_{i_t}[L(y_t,J_t)] - \rE_{i_t}[L(y_t,J^\star_t)]~.
\]
We would like this cumulative regret to be {\em sublinear} in $T$ with high probability (in the random draw of position $y_t$ at the beginning of each episode $t$). Notice that the last term in the RHS of (\ref{e:target}) is independent of $J$, hence $L_M$ will play no role in the regret computation.

\noindent{\bf Algorithm.}
Our ``theoretical" algorithm is described as Algorithm \ref{f:1}. The algorithm operates on an $\epsilon$-cover\footnote
{
Recall that $\cF_\epsilon$ is an $\epsilon$-cover of $\cF$ if $\cF_\epsilon \subseteq\cF$ and for all $x \in \cF$ there is $x' \in \cF_{\epsilon}$ for which $\dist(x,x') \leq \epsilon$. It is easy to see that, given a 2D-scene, the cardinality $|\cF_{\epsilon}|$ of $\cF_\epsilon$ is $O(1/\epsilon^2)$.
}
$\cF_\epsilon$ of $\cF$ and generalized linear probabilities $p_{i_t}(x) = \sigma(\theta^\top \phi(i_t,x))$. The algorithm replaces the above true probabilities in the average path length (\ref{e:target}) with lower confidence estimations $
\sigma\Bigl(\hDelta_t(x) - \epsilon_t(x)\Bigl)
$, 
and then computes $J_t$ by minimizing (\ref{e:target})
over the choice of $k$ points within $\cF_\epsilon$, as well as their order (permutation $\pi$). 
A sequence of signals $
\langle s_{1,t}, \ldots, s_{k'_t,t}\rangle = \langle -1,\ldots, -1, +1\rangle$ observed during episode $t$ is associated with the successful path $J_t$ in that a sequence of $k_t'-1$ negative signals (``$-1$" = object not spotted from $x_{\pi_t(\ell),t}$, for $\ell = 1,\ldots, k'_t-1$) precede a positive signal (``$+1$" = object spotted at $x_{\pi_t(k'_t),t}$). On the other hand, when the object is not found throughout the length-$k$ path, the sequence of signals becomes $\langle s_{1,t}, \ldots, s_{k,t}\rangle = \langle -1,\ldots, -1, -1\rangle$ (this happens with probability $\rP(\cE)$).

When the object is found, Algorithm \ref{f:1} uses the observed signals to update over time a $D$-dimensional weight vector $\htheta$, and a $(D\times D)$-dimensional matrix $M$. Vector $\htheta_t$ is used to estimate $\theta^\top\phi(i_t,x)$, through $\hDelta_t(x)$, while matrix $M_t$ delivers a standard confidence bound via $\epsilon_t^2(x)$. The update rule implements a second-order descent method on logistic loss trying to learn the unknown vector $\theta$ out of the signals $s_{j,t}$. In particular, the update $\htheta_{c_t+j-1} \rightarrow \htheta_{c_t+j}$ is done by computing a standard online Newton step (e.g., \cite{DBLP:journals/ml/HazanAK07}). Notice that at each episode $t$, both matrix $M$ and vector $\htheta
$ get updated $m_t=k'_t$ times, which corresponds to the number of (valid) signals received in that episode. Counter $c_t$ accumulates the number of such updates across (non-failing) episodes. 
On the contrary, when the object is not found, we know that there has been a failure, hence we disregard all (negative) signals received and jump to the next episode with no updates ($m_t = 0$).

From a computational standpoint, calculating $J_t$ as described in Algorithm \ref{f:1} is hard, since the planning problem the algorithm is solving at each episode is essentially equivalent to a {\em Weighted Minimum Latency Problem} (WMLP) \cite{wei2018new}, also called {\em traveling repairman problem}, which, on a generic metric space is NP-hard and also MAX-SNP-hard \cite{b+94}. Fast algorithms are available only for very special metric graphs, like paths \cite{a+86,g+02} edge-unweighted trees \cite{mi89}, trees of diameter 3 \cite{b+94}, trees of constant number of leaves \cite{kpy96}, and the like \cite{wu00}. Even for weighted trees the problem remains NP-hard \cite{s02}. Approximation algorithms are indeed available \cite{c+03}, but they are not practical enough for real-world deployment. In our experiments (Sections \ref{s:proposed_methodology}--\ref{s:experiments}), we implement and compare fast planning approximations to Algorithm \ref{f:1}.

In both the planning and the training of the contextual bandit algorithm, we are using the average path length as a minimization objective because it is easier for the WMLP solver to handle. However, when it comes to evaluating performance in our experiments, we use the Success weighted by Path Length (SPL) metric \cite{anderson2018evaluation} because it is a more established metric in the object-nav literature.

\setlength{\textfloatsep}{0.2cm}
\setlength{\floatsep}{0.2cm}
\begin{algorithm*}[t!]
{\bf Input:} 
Learning rate $\eta >0$, exploration parameter $\alpha \geq 0$, $\epsilon$-cover $\cF_\epsilon$ of $\cF$, $\epsilon > 0$, path length $k$.\\
{\bf Init:} $M_{0} = kI \in \rR^{D\times D}$,\ ${\widehat \theta}_{1} = 0\in \rR^{D}$, $c_1 = 1$.\\
{\bf For $t=1, 2,\ldots, T$}
\begin{enumerate}
%
\item Get object identity $i_t$~, and initial position of the robot $x_{0,t}$~;
\item For $x \in \cF$, set 
\[
{\hDelta_t}(x) = \htheta_{c_t}^\top \phi(i_t,x) \qquad {\mbox{and}}\qquad
\epsilon^2_t(x) =  
\alpha\, \phi(i_t,x)^\top M^{-1}_{c_{t}-1} \phi(i_t,x)
\]
\vspace{-0.15in}
\item Compute $J_t = \langle x_{\pi_t(1),t},\ldots, x_{\pi_t(k),t} \rangle $~ as \hfill\texttt{//solve WMLP at episode $t$}
\[
    J_t = \arg\min_{\stackrel{x_1 \ldots x_k \in \cF_\epsilon}{{\tiny {\mbox{permutation }} }\pi}}\, \sum_{\ell = 1}^k \sigma\Bigl(\hDelta_t(x_{\pi(\ell)}) - \epsilon_t(x_{\pi(\ell)})\Bigl) \sum_{j=1}^\ell\,\dist(x_{\pi(j-1)}, x_{\pi(j)})
\]
\item Observe signal 
$
\begin{cases}
\langle s_{1,t}, \ldots, s_{k'_t,t}\rangle = \langle -1,\ldots, -1, +1\rangle &{\mbox{set $m_t = k'_t$}} \\ {\mbox{or}}\\
\langle s_{1,t}, \ldots, s_{k,t}\rangle = \langle -1,\ldots, -1, -1\rangle &{\mbox{set $m_t = 0$}}
\end{cases}
$
%
\item {\bf For} $j = 1,\ldots, m_t$ (in the order of occurrence of items $x_j$ in $J_t$)
update:
\begin{align*}
M_{c_t+j-1} &= M_{c_t+j-2} + \phi(i_t,x_j) \phi(i_t,x_j)^\top,\\
\htheta_{c_t+j} &= \htheta_{c_t+j-1} + \eta\,\sigma\Bigl(-s_{j,t}\,\htheta_{c_t+j-1}^\top \phi(i_t,x_j) \Bigl)\,s_{j,t}\,M^{-1}_{c_t+j-1} \phi(i_t,x_j)
\end{align*}
%
\vspace{-0.2in}
\item $c_{t+1} \leftarrow c_t + m_t$~.
\end{enumerate}
\caption{Simplified contextual bandit planning algorithm.}\label{f:1}
\end{algorithm*}

\noindent{\bf Regret Analysis.}
From a statistical standpoint, Algorithm \ref{f:1} is a simplified version of the slightly more complex Algorithm \ref{f:2} which is the one our regret analysis applies to. Please refer to Appendix \ref{a:theory} for a discussion on Algorithm \ref{f:2}. 
\begin{theorem}\label{t:main}
Let $D_M = \max_{x,x'\in \cF} \dist(x,x')$ be the diameter of the scene. Also, let the feature mapping $\phi\,:\,[n]\times\rR^d \rightarrow \rR^D$ be such that $||\phi(i,x)||\leq 1$ for all $i \in [n]$ and $x \in \cF$, and let constant $B$ be such that $||\theta||\leq B$.
%
Then a variant of Algorithm \ref{f:1} exists that operates in episode $t$ with an $\epsilon$-covering $\cF_\epsilon$ of $\cF$ with $\epsilon = k/\sqrt{t}$, such that
with probability at least $1-\delta$, with $\delta < 1/e$, the cumulative regret of this algorithm satisfies, on any sequence of objects $i_1,\ldots, i_T$,
\vspace{-0.05in}
\begin{align*}
&R_T(J_1,\ldots,J_T)
= O\Bigl( (1-p)k^2\sqrt{T} + D_M\,k\,\sqrt{(1-p)kT\,\alpha(k,D,T,\delta,B)\,D\log (1+kT)} \Bigl)~,
\end{align*}
where 
$\alpha(k,D,T,\delta,B) =
O\left[e^{2B}\left(k + D\,\log \left(1+\frac{kDT}{\delta}\right)\right)\right]$, and $p = \rP(\cE_t)$ is the (constant) failure probability.
In the above, the big-oh notation hides additive and multiplicative constants independent of $T$, $D$, $B$, $k$, $p$, and $\delta$. 
\end{theorem}
%

In a nutshell, the above analysis provides a high-probability regret guarantee of the form $k^2 D\sqrt{T}$. We present the proof of this theorem in Appendix \ref{a:theory}. The learning rate $\eta$ and the exploration parameter $\alpha$ in Algorithm \ref{f:1} are hyper-parameters.

\section{Proposed Method}\label{s:proposed_methodology}

In this section, we present a practical algorithm that 
replicates Algorithm \ref{f:1}. 
We have three main steps: (a) Sampling $k$ vantage-points that maximally cover the navigable part $\cF$ of the given scene; (b) Importance assignment to the sampled points using an online learning contextual bandit algorithm; and (c) Path planning through the vantage points. We use Model  Predictive  Control  (MPC) \cite{mpc-0,mpc-1, mpc-nav,mpc-nonplanar} to execute the selected path. This guarantees collision-free shortest-path trajectories between vantage points while satisfying given safety constraints, optimality criteria, and kinodynamic models. The resulting system is interpretable and safe.
We will elaborate on each of these.

\vspace{-0.05in}
\subsection{Sampling vantage points}
\label{sec:sampling_vantage_points}
\vspace{-0.04in}
Our algorithm begins by sampling a sparse set of vantage points that are navigable and spread all over the scene in a way that the agent's vision is able to cover the space to a large extent by visiting these points and looking around. First, we extract all the navigable points (at a resolution of $0.1$ meter) from the robot's occupancy map. 
Next, we select our vantage points using the Farthest Point Sub-sampling (FPS) algorithm. 
%
%
%
%
%
%
FPS is a simple, yet effective method of extracting a small number of points from a given point-cloud to cover the extremities of the point-cloud and capture prominent point features. Given a point cloud of size $N$ and a distance metric $d_f$, the FPS algorithm works by picking a starting point (randomly or by some heuristic criterion) from the point cloud and iteratively adding new points that are maximally distant -- according to $d_f$ -- from the current set of selected points, till the required number of points $k$ is reached. This method is popularly used in image processing \cite{eldar1997farthest} and computer vision \cite{qi2017pointnet++}. Since FPS requires $O(N)$ distance calculations in each iteration, we choose the Eucledian distance for $d_f$ in our experiments for computational ease.

\vspace{-0.05in}
\subsection{Node Level Importance Assignment}
\label{sec:node_level_importance_assignment}
\begin{figure}
    \centering
    \includegraphics[width=0.85\linewidth]{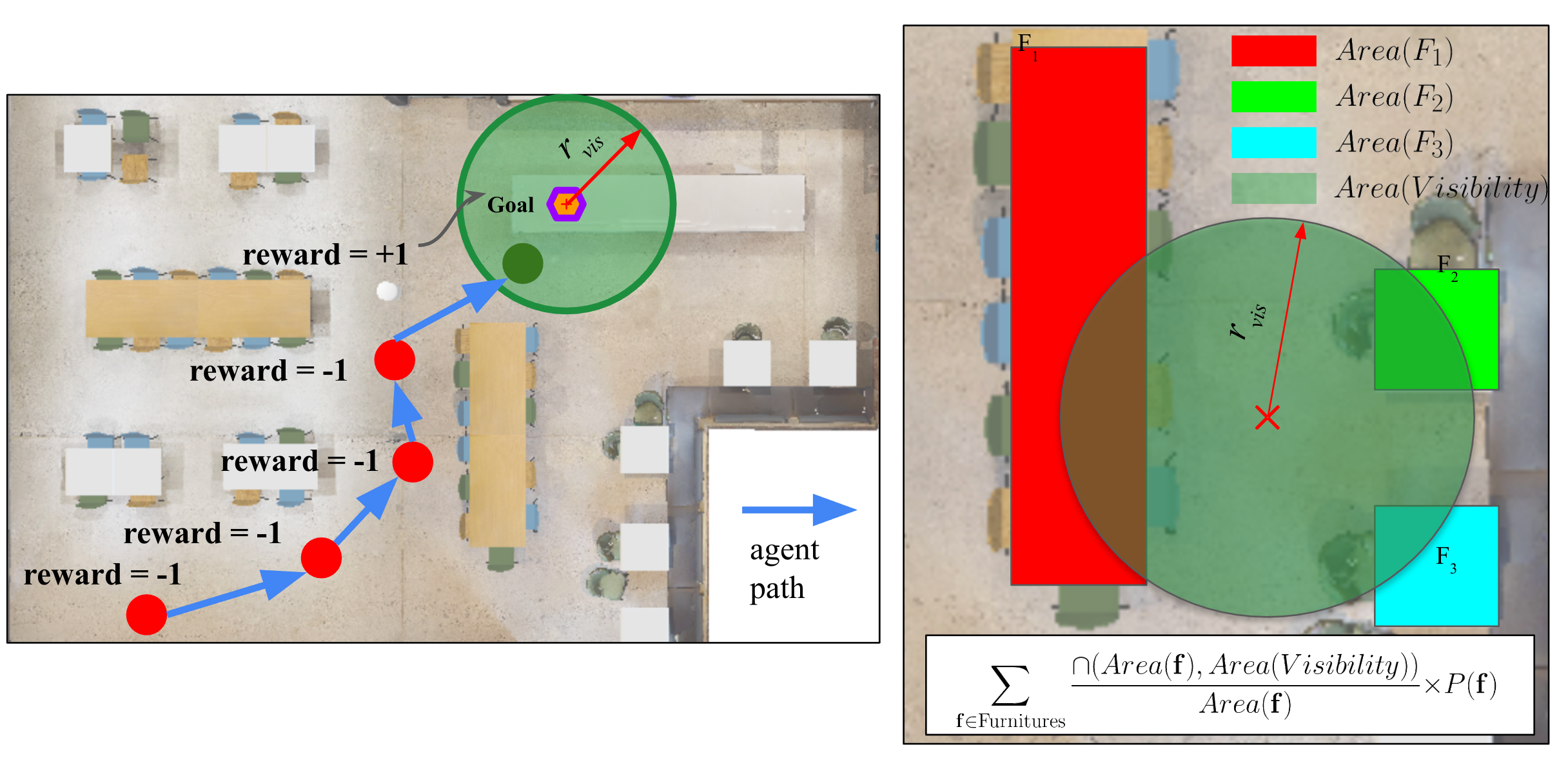}
    \caption{\textbf{Left}: Positive sample augmentation for improved sample efficiency. Radius $r_\mathrm{vis}$ is the robot's visibility range. \textbf{Right}: Calculation of ground-truth likelihood scores. $P(\mathbf{f})$ is the likelihood of the object appearing on furniture $\mathbf{f}$. $\sum_{\mathbf{f}\in \text{Furnitures}} P(\mathbf{f}) = 1$.}
    \label{fig:positive_sample_augmentation}
\end{figure}

\vspace{-0.04in}
Once we have the vantage points, we estimate the importance of each point in the context of the target object. An important vantage point is one that has a high likelihood of the robot spotting the target object standing there. For ease of modelling, we assume that our robot is able to spot the target object equally well from any yaw-angle if it is within its visibility range $r_\mathrm{viz}$. In order to estimate the likelihood of the robot spotting an object (which can be movable) from a given point, we need to explore the environment. For efficient exploration, we formulate the problem as a contextual bandit with vantage point as an arm and follow the principle of \emph{optimism in the face of uncertainty} (e.g., \cite{sutton2018reinforcement}) as described in Algorithm \ref{f:1}. 
Each vantage point is described by a vector with positional, geometric and semantic features of the point along with the identity of the target object. We triangulate the position of the target object once the agent spots it from a vantage point. 
In order to improve learning efficiency, we assign positive training signal (``$+1$") to all the navigable points within $r_\mathrm{vis}$ radius from the object. Figure \ref{fig:positive_sample_augmentation} (left) illustrates this procedure. 

We study two classes of models for estimating the importance of points in our experiments. The first is a generalized linear model (``Gen-Lin'') as presented in Algorithm \ref{f:1} with a minor modification. Since it is difficult for a single generalized linear function to model multiple potentially non-linear decision boundaries corresponding to different object classes, we use a disjoint set of model parameters for each object class. The second is a simple two-layer fully connected neural network model, approximated via Neural Tangent Kernels (NTK), as contained in Algorithms 1-2 in  \cite{zhou2020neural} (``Neural") for computing uncertainties $\epsilon_t(x)$ (Step 2 in Algorithm \ref{f:1}) and updating $\widehat\theta$ (Step 5 in Algorithm \ref{f:1}). Unlike the ``GenLin'' model, the same neural network parameters are shared across all classes of objects.

\vspace{-0.05in}
\subsection{Planning}
\vspace{-0.04in}
After estimating the importance-scores of the vantage points, we derive their sequence of visitation by representing them as a graph -- where each node is a vantage point and the edges contain the $A^\star$ distance between vantage points -- and solving the Weighted Minimum Latency Problem (WMLP) (e.g., \cite{b+94,wei2018new}). WMLP tries to minimize the average waiting time of each node in a graph, weighted by its importance score being reached by a travelling agent. 
In our case, the solution of the WMLP minimizes the average distance traveled by the robot to reach the target object -- the average being over the position of the object and the initial position of the robot. We consider two different ways of solving the WMLP in our setting.

\noindent{\bf CP-SAT.}
The first approach directly faces the underlying optimization problem. We relied on a satisfiability (SAT)-based constraint programming (CP) solver \cite{solver-book} from Google OR-Tools \cite{gdev_ortools} that uses a lazy clause generation solver on top of a SAT solver to reach its solution conditioned on vantage-points, starting position, and predicted relevance of the points. Although this approach is direct and principled, the running time of this solver may increase drastically as the number of points grows.

\noindent{\bf One-step greedy.} Starting at $x$, the next point $x'$ is chosen to maximize a weighted combination of the estimated likelihood $\hat{p}_{i}(x')$ of spotting the target object $i$, and the inverse of the $A^\star$ distance traveled to reach it:
\begin{equation}
       x' = \argmax_{x''\in \text{\{unvisited\ points\}}}  \frac{\alpha_p}{\dist(x,x'')} + (1-\alpha_p)\,\hat{p}_{i}(x'')~,
\end{equation}
where $\alpha_p \in [0,1]$ is a hyper-parameter whose value should be chosen to achieve a good trade-off between minimizing the traveled distance in the next step and maximizing the likelihood of spotting the target object. The case of $\alpha_p=0$ corresponds to greedily choosing the unvisited point with the highest estimated likelihood, while the case of $\alpha_p=1$ would greedily choose the closest unvisited point. In the above, $\hat{p}_{i}(x)$ will be computed as suggested by Algorithm \ref{f:1} via a lower confidence scheme of the form
$\hat{p}_{i}(x) = \sigma\Bigl(\hDelta(x) - \epsilon(x)\Bigl)$.
The one-step greedy approach is myopic, as it greedily optimizes for the next step only, but is also much faster to run, hence it should be interpreted here as a fast approximation to the CP-SAT solution.

\section{Experiments}\label{s:experiments}
\vspace{-0.14in}
We run experiments in two simulated and one real office kitchen environments. The two simulated kitchens have areas $80$ sq.m. and $120$ sq.m., respectively. Each experiment involves training the agent for $200$ episodes followed by evaluation with frozen parameters in the same environment. For the real kitchen experiment, we train the bandit model on a snapshot of the map in our simulator and evaluate in the real environment.

\begin{figure}
    \centering
    \includegraphics[width=0.9\linewidth]{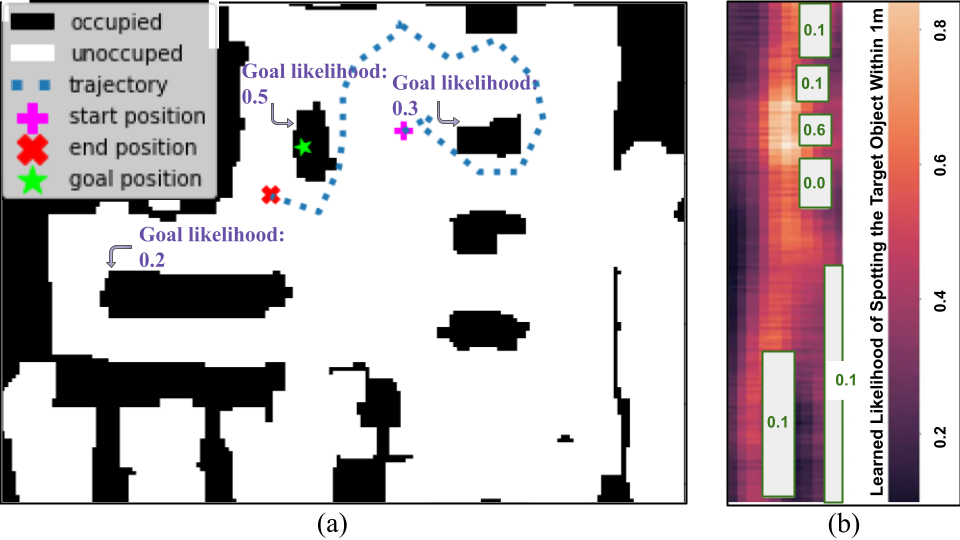}
    \vspace{-0.07in}
    \caption{{\small {\bf (a)} Real-kitchen sample trajectory for CP-SAT planner with the Neural model. {\bf (b)} Heat map of the estimated likelihood of spotting the goal object (``bottle'') within a distance of $1$m along with ground truth likelihoods (in green) of the object occurring on the surface of each furniture.}}
    \label{fig:real-trajectory-sim-heatmap}
\end{figure}

The simulated environments have photo-realistic scenes generated from Matterport scans \cite{chang2017matterport3d} and Bullet \cite{coumans2019} based physics simulation. Our robot is a differential-drive wheeled robot from Everyday Robots \footnote{\url{https://everydayrobots.com/}\label{robot-company}}, which has a 3D LiDAR in the front, and depth sensors mounted on its head. It is capable of accurate localization and safe point-to-point navigation. We consider two different ways of solving the WMLP in our setting -- a) directly solving the optimization problem using CP-SAT, a satisfiability (SAT)-based constraint programming (CP) solver \cite{solver-book} from Google OR-Tools \cite{gdev_ortools}; and b) a one-step greedy approach that maximizes a weighted combination of the estimated likelihood of spotting the target object and the inverse of the $A^\star$ distance traveled to reach it -- to choose the next vantage point in the visitation sequence. Although myopic, the latter is a faster approximation of the CP-SAT algorithm that despite being direct and more principled, suffers from drastic increases in running time with growing number of vantage points (See Figure \ref{fig:cpsat-running-time}). We use Model Predictive Control \cite{francis2022learning, camacho2013model} to execute a path. Each vantage point $x$ is described by a feature-vector $\phi(i, x)$ consisting of a one-hot encoding of the target object $i$ and a flattened $16\times 16$ patch of the wall-distance map centered at the point $x$ (see also Figure \ref{fig:wall_dist_ft} in Appendix \ref{object-dist}). A sinusoidal positional encoding \cite{vaswani2017attention} vector is appended to represent the location of the point in the map.  Map-resolution and positional encoding dimension are hyper-parameters. The normalizer for the feature-vector is also a hyper-parameter that is chosen from among: a) zero mean, unit standard-deviation, and b) unit $l^2$-norm. All hyper-parameters ($\eta$ and $\alpha$ for Algorithm \ref{f:1}, $\alpha_p$ for One-step greedy, the learning rate and batch size in Algorithms 1-2 in \cite{zhou2020neural} for Neural, the positional embededing size and the feature vector normalization for mapping $\phi(i,x)$, and the sigmoidal scale $s$ for the sigmoid in Algorithm \ref{f:1}) are tuned across suitable ranges (Table \ref{tab:hyp-range}) using a Gaussian-Process Bandit based Blackbox optimizer \cite{Vizier} to maximize \textbf{success weighted by path length (SPL)} \cite{objectnaveval} over the training episodes (Appendix~\ref{hyperparameters}).

For training in simulation, we consider the agent has successfully reached its goal if and when it visits a vantage point that is within $r_\mathrm{vis}$ radius from the target object. For learning good quality likelihood maps, we set $r_\mathrm{vis}=1$m during training although the default value of $r_\mathrm{vis}=2.5$m is used during evaluation in the simulated environments. For success during evaluation in the real environment, the robot must visually detect the target object and drive up to a grasping range of the object. For object detection, we use our implementation of the ViLD detector \cite{gu2021open} that has a true positive rate of $84.6\%$ across our test objects.
%
We have five categories of target objects: ``bottle'', ``can'', ``cup'', ``bowl'' and ``chips-bag'' each of which has the same frequency of occurrence across episodes and in any given episode we have a single instance of the target object in the environment.
We compare the performances of our proposed framework using the generalized linear (``Gen-Lin'') and neural (``Neural'') models for training the contextual bandit agent and ``CP-SAT'' and one-step greedy (``Greedy'') algorithms for path planning to a purely geometric approach that solves the Travelling Salesman Problem \cite{lenstra1975some} (``TSP''). We assign a time budget of $30$ seconds to the CP-SAT solver to have a realistic bound on robot response time. 
Each algorithm is evaluated over $50$ episodes in real and $300$ episodes in simulated environments. During training, whenever Algorithm \ref{f:1} receives a positive signal on a given vantage point, this signal is extended to nearby points (see Appendix \ref{sec:modelling} for details).
Figure \ref{fig:real-trajectory-sim-heatmap} (a) shows a sample trajectory during the real kitchen evaluation. Figure \ref{fig:real-trajectory-sim-heatmap} (b) shows a sample learned likelihood map for the simulated kitchen-1 environment. 


We compare the performance of the agents on the following metrics: a) rate of success during evaluation (``Eval. Succ.''), b) SPL \cite{objectnaveval} during training (``Train SPL''), and c) SPL during evaluation (``Eval. SPL''). Higher ``Train SPL'' indicates faster rate of convergence. Table \ref{table:all_results} presents the results of our first study, where we compare different learning and planning approaches for an arbitrary spatial distribution of test objects. The columns labeled ``GT-Scores'' use ground truth likelihoods of the vantage points computed as shown in Figure \ref{fig:positive_sample_augmentation} (right) and mapped in Figures \ref{fig:gt_obj_dist_kitchen_sim_1}, and \ref{fig:gt_obj_dist_kitchen_sim_2} in Appendix \ref{object-dist}. 
\renewcommand{\tabcolsep}{1pt}
\begin{table*}[]
\centering
\begin{tiny}
\begin{tabular}{c|ccc|ccccccc|ccccccc|}
\cline{2-18}
 &
  \multicolumn{3}{c|}{\textbf{Real Kitchen Env.}} &
  \multicolumn{7}{c|}{\textbf{Kitchen Environment 1}} &
  \multicolumn{7}{c|}{\textbf{Kitchen Environment 2}} \\ \cline{2-18} 
 &
  \multicolumn{1}{c|}{} &
  \multicolumn{2}{c|}{\textbf{Neural}} &
  \multicolumn{1}{c|}{} &
  \multicolumn{2}{c|}{\textbf{Gen-Lin}} &
  \multicolumn{2}{c|}{\textbf{Neural}} &
  \multicolumn{2}{c|}{\cellcolor[HTML]{C0C0C0}\textbf{GT-Scores}} &
  \multicolumn{1}{c|}{} &
  \multicolumn{2}{c|}{\textbf{Gen-Lin}} &
  \multicolumn{2}{c|}{\textbf{Neural}} &
  \multicolumn{2}{c|}{\cellcolor[HTML]{C0C0C0}\textbf{GT-Scores}} \\ \cline{3-4} \cline{6-11} \cline{13-18} 
 &
  \multicolumn{1}{c|}{\multirow{-2}{*}{\textbf{TSP}}} &
  \multicolumn{1}{c|}{\textbf{Greedy}} &
  \textbf{CP-SAT} &
  \multicolumn{1}{c|}{\multirow{-2}{*}{\textbf{TSP}}} &
  \multicolumn{1}{c|}{\textbf{Greedy}} &
  \multicolumn{1}{c|}{\textbf{CP-SAT}} &
  \multicolumn{1}{c|}{\textbf{Greedy}} &
  \multicolumn{1}{c|}{\textbf{CP-SAT}} &
  \multicolumn{1}{c|}{\cellcolor[HTML]{C0C0C0}\textbf{Greedy}} &
  \cellcolor[HTML]{C0C0C0}\textbf{CP-SAT} &
  \multicolumn{1}{c|}{\multirow{-2}{*}{\textbf{TSP}}} &
  \multicolumn{1}{c|}{\textbf{Greedy}} &
  \multicolumn{1}{c|}{\textbf{CP-SAT}} &
  \multicolumn{1}{c|}{\textbf{Greedy}} &
  \multicolumn{1}{c|}{\textbf{CP-SAT}} &
  \multicolumn{1}{c|}{\cellcolor[HTML]{C0C0C0}\textbf{Greedy}} &
  \cellcolor[HTML]{C0C0C0}\textbf{CP-SAT} \\ \hline
\multicolumn{1}{|l|}{\textbf{Train SPL}} &
  \multicolumn{1}{c|}{-} &
  \multicolumn{1}{c|}{-} &
  - &
  \multicolumn{1}{c|}{-} &
  \multicolumn{1}{c|}{$0.32$} &
  \multicolumn{1}{c|}{$0.31$} &
  \multicolumn{1}{c|}{$0.30$} &
  \multicolumn{1}{c|}{$0.27$} &
  \multicolumn{1}{c|}{\cellcolor[HTML]{C0C0C0}-} &
  \cellcolor[HTML]{C0C0C0}- &
  \multicolumn{1}{c|}{-} &
  \multicolumn{1}{c|}{$0.26$} &
  \multicolumn{1}{c|}{$0.27$} &
  \multicolumn{1}{c|}{$0.23$} &
  \multicolumn{1}{c|}{$0.13$} &
  \multicolumn{1}{c|}{\cellcolor[HTML]{C0C0C0}-} &
  \cellcolor[HTML]{C0C0C0}- \\ \hline
\multicolumn{1}{|c|}{\textbf{Eval. Succ.}} &
  \multicolumn{1}{c|}{$0.88$} &
  \multicolumn{1}{c|}{$0.80$} &
  $0.92$ &
  \multicolumn{1}{c|}{$0.89$} &
  \multicolumn{1}{c|}{$0.88$} &
  \multicolumn{1}{c|}{$0.89$} &
  \multicolumn{1}{c|}{$0.90$} &
  \multicolumn{1}{c|}{$0.90$} &
  \multicolumn{1}{c|}{\cellcolor[HTML]{C0C0C0}$0.90$} &
  \cellcolor[HTML]{C0C0C0}$0.90$ &
  \multicolumn{1}{c|}{$0.56$} &
  \multicolumn{1}{c|}{$0.80$} &
  \multicolumn{1}{c|}{$0.78$} &
  \multicolumn{1}{c|}{$0.74$} &
  \multicolumn{1}{c|}{$0.67$} &
  \multicolumn{1}{c|}{\cellcolor[HTML]{C0C0C0}$0.82$} &
  \cellcolor[HTML]{C0C0C0}$0.80$ \\ \hline
\multicolumn{1}{|l|}{\textbf{Eval. SPL}} &
  \multicolumn{1}{c|}{$0.37$} &
  \multicolumn{1}{c|}{$0.38$} &
  $0.42$ &
  \multicolumn{1}{c|}{$0.38$} &
  \multicolumn{1}{c|}{$0.42$} &
  \multicolumn{1}{c|}{$0.47$} &
  \multicolumn{1}{c|}{$0.40$} &
  \multicolumn{1}{c|}{$0.39$} &
  \multicolumn{1}{c|}{\cellcolor[HTML]{C0C0C0}$0.43$} &
  \cellcolor[HTML]{C0C0C0}$0.51$ &
  \multicolumn{1}{c|}{$0.22$} &
  \multicolumn{1}{c|}{$0.26$} &
  \multicolumn{1}{c|}{$0.29$} &
  \multicolumn{1}{c|}{$0.25$} &
  \multicolumn{1}{c|}{$0.20$} &
  \multicolumn{1}{c|}{\cellcolor[HTML]{C0C0C0}$0.33$} &
  \cellcolor[HTML]{C0C0C0}$0.34$ \\ \hline
\end{tabular}
\end{tiny}
\caption{Empirical evaluation of our agents on $3$ environments in terms of object-nav metrics.}
\label{table:all_results}
\end{table*}
%
%
%
%
%
For both training and evaluation, we use $25$ vantage points for the real environment and the kitchen environment 1 and $50$ for kitchen environment 2. 

Our first observation is the significant improvement in performance achieved by our planners using ``GT-Scores'' over ``TSP'' in all the environments, and this validates the importance of estimating the importance of the vantage points in the context of the target object in addition to optimizing for the room geometry. The performances for ``GT-Scores'' provide an upper bound for the agents that learn the likelihood function through exploration. Although the performance of ``CP-SAT'' shines in the real evaluation, under the planning time budget of $30$ seconds, the performance of our proposed one-step greedy solver regularly matches up and often beats the CP-SAT solver, especially in larger and more cluttered kitchen environment 2. The performance of the ``Neural'' model often seems to lag behind the ``Gen-Lin'' model. This is because the ``Gen-Lin'' model does not have to generalize across all the object categories with the same set of parameters and hence has a less challenging learning problem to solve. Please visit the project website\footref{website} for videos of real world tests. 

\section{Limitations and Future Work}\label{s:limitations}
%









Our proposed approach can get adversely affected due to: 1) detection failure, 2) slowness of WMLP, and 3) early stopping of CP-SAT solver (not running the CP-SAT solver until the end may give us feasible but poor solutions). Inclusion of orientation along with the robot's base position can help in mitigating missed detections. Using richer feature embeddings can also improve object detection from distance. Related to that, the choice of the network architecture in the Neural model is severely limited by our usage of the NTK approximation to compute confidence bounds. Leveraging more time-efficient approximation schemes may allow for more complex (and potentially more accurate) network architectures. Faster convergence is possible in training by using a likelihood-guided sampling scheme but this may also create opportunities for local minima. These are among the missing aspects we are currently investigating.

\newpage
\bibliographystyle{plain}
\bibliography{arxiv_neurips.bib}

\newpage
\newpage
\appendix
\counterwithin{figure}{section}
\counterwithin{table}{section}
\section*{APPENDIX}

\section{Theoretical Underpinning}\label{a:theory}

In this section, we present Algorithm \ref{f:2} and prove Theorem \ref{t:main} from the main body of the paper. We first need a couple of ancillary lemmas.

\begin{lemma}\label{l:onestep}
Let Algorithm \ref{f:2} be run on an $\epsilon$-cover $\cF_\epsilon$ with $\epsilon = O(1/\sqrt{t})$. 
Let episode $t$ be such that 
$|\theta^\top \phi(i_t,x) - \phi(i_t,x)^\top \htheta'_{c_t}| \leq \epsilon_t(x)$ for all $x \in \cF$. Also, let $D_M = \max_{x,x'\in \cF} \dist(x,x')$ denote the diameter of the scene. Then
\begin{align*}
&\rE_{i_t}[L(y_t,J_t)] - \rE_{i_t}[L(y_t,J^\star)] = O\left((1-\rP(\cE_t))\left(\frac{k^2}{\sqrt{t}}+  D_M\,k\, \sum_{\ell = 1}^k \epsilon_t(x_{\pi_t(\ell),t})\right)\right) ~,
\end{align*}
where $\cE_t$ denotes the failure event during episode $t$.
\end{lemma}
{\em Proof.}
We fix episode $t$ and remove subscript $t$ and $c_t$ for convenience. As short-hands, let us denote by $J = \langle x_1,\ldots, x_k\rangle$ the path computed by Algorithm \ref{f:2} in episode $t$ and by $J^\star_\epsilon = \langle x^\star_{1,\epsilon},\ldots, x^\star_{k,\epsilon}\rangle$ the minimizer of (\ref{e:target}) when the $k$ vantage points are constrained to lie in the $\epsilon$-cover $\cF_\epsilon$. We clearly have
\begin{align*}
|\rE_i[L(y,J^\star_\epsilon)] - \rE_i[L(y,J^\star)]| 
&\leq (1-\rP(\cE_t))k(k+1)\epsilon\\ 
&= O\left((1-\rP(\cE_t))\frac{k^2}{\sqrt{t}} \right) ~.
\end{align*}
Moreover,
\begin{align*}
&\frac{1}{(1-\rP(\cE_t))}\,\rE_i[L(y,J)] - \rE_i[L(y,J^\star_\epsilon)]\\ 
&=\sum_{\ell = 1}^k p_{i}(x_{\ell})\sum_{j=1}^\ell\,\dist(x_{j-1}, x_{j}) 
- \sum_{\ell = 1}^k p_{i}(x^\star_{\ell,\epsilon})\sum_{j=1}^\ell\,\dist(x^\star_{j-1,\epsilon}, x^\star_{j,\epsilon})\\
&\leq
\sum_{\ell = 1}^k \sigma\Bigl(\theta^\top \phi(i,x_{\ell})\Bigl)\sum_{j=1}^\ell\,\dist(x_{j-1}, x_{j})
-\sum_{\ell = 1}^k 
\sigma\Bigl(\htheta^\top \phi(i,x^\star_{\ell,\epsilon})-\epsilon(x^\star_{\ell,\epsilon})\Bigl)
\sum_{j=1}^\ell\,\dist(x^\star_{j-1,\epsilon}, x^\star_{j,\epsilon})~.
\end{align*}
In turn, the above is upper bounded by
\begin{align*}
&\sum_{\ell = 1}^k \sigma\Bigl(\theta^\top \phi(i,x_{\ell})\Bigl)\sum_{j=1}^\ell\,\dist(x_{j-1}, x_{j})
-\sum_{\ell = 1}^k 
\sigma\Bigl(\htheta^\top \phi(i,x_{\ell})-\epsilon(x_{\ell})\Bigl)
\sum_{j=1}^\ell\,\dist(x_{j-1}, x_{j})\\
&\leq
\sum_{\ell = 1}^k \sigma\Bigl(\htheta^\top \phi(i,x_{\ell})+\epsilon(x_{\ell})\Bigl)\sum_{j=1}^\ell\,\dist(x_{j-1}, x_{j})
- \sum_{\ell = 1}^k 
\sigma\Bigl(\htheta^\top \phi(i,x_{\ell})-\epsilon(x_{\ell})\Bigl)
\sum_{j=1}^\ell\,\dist(x_{j-1}, x_{j})\\
&\leq
\sum_{\ell = 1}^k 2\epsilon(x_{\ell})\sum_{j=1}^\ell\,\dist(x_{j-1}, x_{j}) \\
&=
O\left(D_M\,k\,\sum_{\ell = 1}^k \epsilon(x_{\ell}) \right)~.
\end{align*}
Putting together proves the claim.

\begin{lemma}\label{l:mahalanobis}
Let $B >0$ be such that $\theta^\top \phi(i,x) \in [-B,B]$ for all $i \in [n]$ and $x \in \cF$. Moreover, 
let $c_{\sigma}$ and $c_{\sigma'}$ be two positive constants such that, for all $\Delta \in [-D,D]$ the conditions
$0 < 1-c_{\sigma} \leq \sigma(\Delta) \leq c_{\sigma} < 1$ and $\sigma'(\Delta) \geq c_{\sigma'}$ hold.
Then with probability at least $1-\delta$, with $\delta < 1/e$, we have
\[
d_{c_t-1}(\theta,\htheta'_{c_t})
\leq
\alpha(k,D,T,\delta,B)~,
\]
uniformly over $c_t \in [kT]$, where
\begin{align*}
&\alpha(k,D,T,\delta,B)\\ 
&=
O\Biggl(kB^2 
+ \left(\frac{c_{\sigma}}{c_{\sigma'}}\right)^2 D\log \left(1+ \frac{1}{k}\Bigl(\frac{t\,c_{\sigma}}{1-c_{\sigma}} + \log\frac{t+1}{\delta}\Bigl) \right)
+ \left(\left(\frac{c_{\sigma}}{c_{\sigma'}}\right)^2 + \frac{1+B}{c_{\sigma'}} \right) \log \frac{k(t+1)}{\delta}\Biggl)~.
\end{align*}
\end{lemma}
{\em Proof.} 
The proof follows from standard concentration arguments applied to the logistic loss, which Algorithm \ref{f:2} implicitly operates on. See, e.g., \cite{pmlr-v151-santara22a}, Lemma 5 therein which, in turn, relies on \cite{DBLP:journals/ml/HazanAK07} and \cite{go12}. The argument therein can be applied to the non-failing episodes, that is, those episodes on which state updates occur. In our bound above we are simply over-approximating the number of non-failing episodes within the first $t$ episodes with $t$ itself.

%
\vspace{0.25in}
{\em Proof of Theorem \ref{t:main}}
From Lemma \ref{l:mahalanobis} and the Cauchy-Schwarz inequality it follows that 
\begin{align*}
(\theta^\top \phi(i,x) - \phi(i,x)^\top \htheta'_{c_t})^2 
&\leq \phi(i,x)^\top M_{c_t-1}^{-1}\phi(i,x)\,  d_{c_t-1}(\theta,\htheta'_{c_t})\\
&\leq \left(\phi(i,x)^\top M_{c_t-1}^{-1}\phi(i,x)\right)\,\alpha(k,D,T,\delta,B)
\end{align*}
for all $i \in [n]$ and $x \in \cF$. Hence we can apply Lemma \ref{l:onestep} with 
\[
\epsilon^2_t(x) = \left(\phi(i,x)^\top M_{c_t-1}^{-1}\phi(i,x)\right)\,\alpha(k,D,T,\delta,B)~.
\]
Let $\cE_t$ denote the failure event at episode $t$, with $\rP(\cE_t) = p$ for all $t$. Summing over $t= 1,\ldots, T$, we can write
\begin{align*}
&\sum_{t=1}^T \Bigl(\rE_{i_t}[L(y_t,J_t)] - \rE_{i_t}[L(y_t,J^\star)] \Bigl) \\
&=
O\left((1-p)k^2\sqrt{T} + D_M\,k\,\rE\left[\sum_{t=1,\,\cE_t=0}^T \sum_{\ell=1}^k \epsilon_t(x_{\pi_t(\ell),t})\right]\right)
= O\Biggl((1-p)k^2\sqrt{T}+ D_M\,k\,\sqrt{\alpha(k,D,T,\delta,B)}\,\,\rE 
\Biggl)~,
\end{align*}
where $\rE$ is a short-hand for
\[
\rE\Biggl[\sum_{t=1,\,\cE_t=0}^T  \sum_{\ell=1}^k 
\sqrt{\phi(i_t,x_{\pi_t(\ell),t})^\top M_{c_t-1}^{-1}\phi(i_t,x_{\pi_t(\ell),t})}\Biggl]~.
\]
We now follow similar arguments as in the proof of Theorem 1 in \cite{pmlr-v151-santara22a} by focusing on 
\[
\sum_{t=1,\,\cE_t=0}^T \sum_{\ell=1}^k 
\phi(i_t,x_{\pi_t(\ell),t})^\top M_{c_t-1}^{-1}\phi(i_t,x_{\pi_t(\ell),t})~.
\]
First, by virtue of Lemma 6 in \cite{pmlr-v151-santara22a}, we have, for each $t$,
\begin{align*}
\sum_{\ell=1}^k 
& \phi(i_t,x_{\pi_t(\ell),t})^\top M_{c_t-1}^{-1}\phi(i_t,x_{\pi_t(\ell),t})
\leq
e\,\sum_{\ell=1}^k 
\phi(i_t,x_{\pi_t(\ell),t})^\top M_{c_t-1+\ell}^{-1}\,\phi(i_t,x_{\pi_t(\ell),t})~,
\end{align*}
so that
\begin{align*}
&\sum_{t=1,\,\cE_t=0}^T \sum_{\ell=1}^k 
\phi(i_t,x_{\pi_t(\ell),t})^\top M_{c_t-1}^{-1}\phi(i_t,x_{\pi_t(\ell),t})
\leq 
e\,\sum_{t=1,\,\cE_t=0}^T \sum_{\ell=1}^k 
\phi(i_t,x_{\pi_t(\ell),t})^\top M_{c_t-1+\ell}^{-1}\,\phi(i_t,x_{\pi_t(\ell),t}) \\
&\qquad= O\left(D\log (1+kT) \right)~,
\end{align*}
the last inequality following from standard upper bounds
(e.g., \cite{cb+05,APS-2011}). As a consequence
\begin{align*}
\sum_{t=1,\,\cE_t=0}^T &\sum_{\ell=1}^k 
\sqrt{\phi(i_t,x_{\pi_t(\ell),t})^\top M_{c_t-1}^{-1}\phi(i_t,x_{\pi_t(\ell),t})}
= O\left(\sqrt{kD\log (1+kT)\sum_{t=1}^T (1-\cE_t) }\right)~,
\end{align*}
which we plug back. Using the concavity of the square root, this allows us to obtain
\begin{align*}
\sum_{t=1}^T \Bigl(\rE_{i_t}[L(y_t,J_t)] &- \rE_{i_t}[L(y_t,J^\star)] \Bigl)\\
&= O\Bigl( (1-p)k^2\sqrt{T} + D_M\,k\,\sqrt{(1-p)kT\,\alpha(k,D,T,\delta,B)\,D\log (1+kT)} \Bigl)~.\\
\end{align*}
Finally, observe that, since $\sigma(z) =\frac{\exp{(z)}}{1+\exp{(z)}}$, we have within the expression for $\alpha(k,D,T,\delta,B)$ in Lemma \ref{l:mahalanobis}, $c_{\sigma} = \frac{e^B}{1+e^B}$ (hence $\frac{c_{\sigma}}{1-c_{\sigma}} = e^B$), and $c_{\sigma'} = e^{-B}/(1+e^{-B})^2 \geq e^{-B}/4$. Plugging back concludes the proof.

%

%

In a nutshell, the above analysis provides a high-probability regret guarantee of the form $k^2 D\sqrt{T}$, when hyperparameters $\eta$ and $\alpha$ in Algorithm \ref{f:1} are assigned specific values, as detailed in Algorithm \ref{f:2}.

\setlength{\textfloatsep}{0.2cm}
\setlength{\floatsep}{0.2cm}
\begin{algorithm*}
{\bf Input:} 
$\epsilon$-cover $\cF_\epsilon$ of $\cF$, $\epsilon > 0$, path length $k$, maximal range $B>0$.\\
{\bf Init:} $M_{0} = kI \in \rR^{D\times D}$,\ ${\widehat \theta}_{1} = 0\in \rR^{D}$, $c_1 = 1$.\\
{\bf For $t=1, 2,\ldots, T$}
\begin{enumerate}
%
\item Get object identity $i_t$~, and initial position of the robot $x_{0,t}$~;
\item For $x \in \cF$, set 
\[
{\hDelta_t}(x) = \phi(i_t,x)^\top \htheta'_{c_t}(x) \qquad {\mbox{and}}\qquad
\epsilon^2_t(x) =  
\alpha(k,D,T,\delta,B)\, \phi(i_t,x)^\top M^{-1}_{c_{t}-1} \phi(i_t,x)~,
\]
where
\begin{align*}
\htheta'_{c_t}(x) = \arg\min_{\theta\,:\,-B \leq \theta^\top \phi(i_t,x) \leq B} d_{c_t-1}(\theta,\htheta_{c_t})~;
\end{align*}
and
\begin{align*}
\alpha(k,D,T,\delta,B) 
&= 
O\Biggl(kB^2 
+ \left(\frac{c_{\sigma}}{c_{\sigma'}}\right)^2 D\log \left(1+ \frac{1}{k}\Bigl(\frac{t\,c_{\sigma}}{1-c_{\sigma}} + \log\frac{t+1}{\delta}\Bigl) \right)\\
&\qquad\qquad+ \left(\left(\frac{c_{\sigma}}{c_{\sigma'}}\right)^2 + \frac{1+B}{c_{\sigma'}} \right) \log \frac{k(t+1)}{\delta}\Biggl)
\end{align*}
\item Compute $J_t = \langle x_{\pi_t(1),t},\ldots, x_{\pi_t(k),t} \rangle $~ as \hfill\texttt{//solve WMLP at episode $t$}
\[
    J_t = \arg\min_{\stackrel{x_1 \ldots x_k \in \cF_\epsilon}{{\tiny {\mbox{permutation }} }\pi}}\, \sum_{\ell = 1}^k \sigma\Bigl(\hDelta_t(x_{\pi(\ell)}) + \epsilon_t(x_{\pi(\ell)})\Bigl) \sum_{j=1}^\ell\,\dist(x_{\pi(j-1)}, x_{\pi(j)})
\]
\item Observe signal 
$
\begin{cases}
\langle s_{1,t}, \ldots, s_{k'_t,t}\rangle = \langle -1,\ldots, -1, +1\rangle &{\mbox{set $m_t = k'_t$}} \\ {\mbox{or}}\\
\langle s_{1,t}, \ldots, s_{k,t}\rangle = \langle -1,\ldots, -1, -1\rangle &{\mbox{set $m_t = 0$}}
\end{cases}
$
%
\item {\bf For} $j = 1,\ldots, m_t$ (in the order of occurrence of items $x_j$ in $J_t$)
update:
\begin{align*}
M_{c_t+j-1} &= M_{c_t+j-2} + \phi(i_t,x_j) \phi(i_t,x_j)^\top,\\
\htheta_{c_t+j} & = \htheta'_{c_t+j-1} + \frac{1}{c_{\sigma'}} M^{-1}_{c_t+j-1}\nabla_{j,t}~,
\end{align*}
where
\(
\nabla_{j,t} 
             = \sigma(-s_{j,t}\,\hDelta'_{t}(x_j))\,s_{j,t}\,\phi(i_t,x_j)~,
\)
where $\hDelta'_{t}(x_j) = \phi(i_t,x_j)^\top \htheta'_{c_t+j-1}$\\ 
with
$$
\htheta'_{c_t+j-1} = \arg\min_{\theta\,:\,-B \leq \theta^\top \phi(i_t,x_j) \leq B} d_{c_t+j-2}(\theta,\htheta_{c_t+j-1})~;
$$
%
\item $c_{t+1} \leftarrow c_t + m_t$~.
\end{enumerate}
\caption{\label{f:2}Contextual bandit planning algorithm.}
\end{algorithm*}


\section{Data augmentation}\label{sec:modelling}

Each vantage point is described by a vector with positional, geometric and semantic features of the point along with the identity of the target object. We triangulate the position of the target object once the agent spots it from a vantage point. 
In order to improve learning efficiency, we assign positive training signal (``$+1$'') to all the navigable points within $r_\mathrm{vis}$ radius from the object. Figure \ref{fig:positive_sample_augmentation} (left) illustrates this procedure.

\section{Object distributions} \label{object-dist}

\begin{figure}
    \centering
    \includegraphics[width=\textwidth]{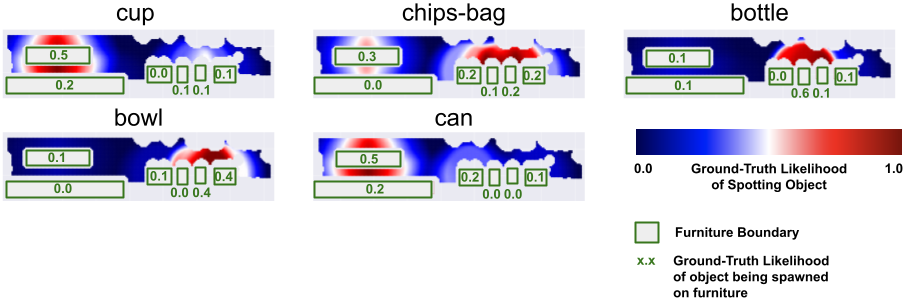}
    \caption{Map of Simulated Kitchen 1 with distribution of occurrence of the five target objects categories ``cup'', ``chips-bag'', ``bottle'', ``bowl'', and ``can''.}
    \label{fig:gt_obj_dist_kitchen_sim_1}
    \vspace{0.2in}
\end{figure}

\begin{figure}
    \centering
    \includegraphics[width=\textwidth]{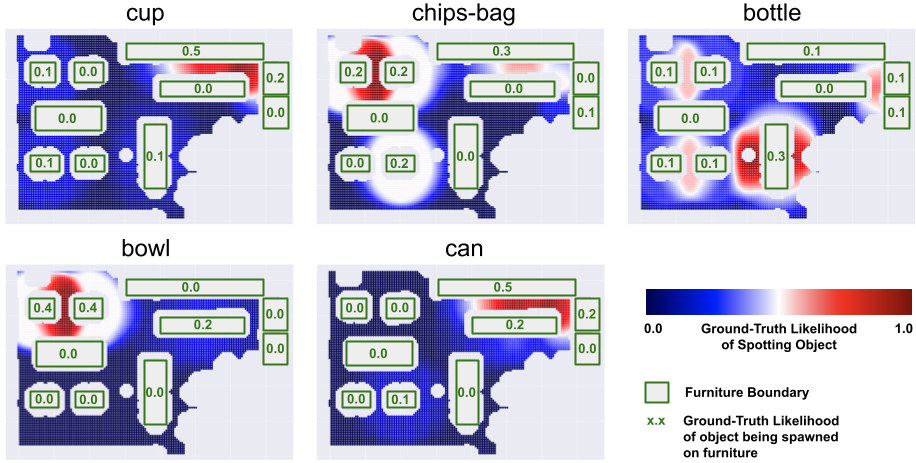}
    \caption{Map of Simulated Kitchen 2 with distribution of occurrence of the five target objects categories ``cup'', ``chips-bag'', ``bottle'', ``bowl'', and ``can''.}
    \label{fig:gt_obj_dist_kitchen_sim_2}
\end{figure}



In this section, we describe the way objects are spawned in the environment in simulation. We only consider objects that are kept on table-tops. As shown in Figures \ref{fig:gt_obj_dist_kitchen_sim_1} and \ref{fig:gt_obj_dist_kitchen_sim_2}, each table in the environment has a certain probability of housing the object. In each episode, for each object category, a table is sampled from the corresponding probability distribution. A location on the surface of the selected table is then picked uniformly at random to determine the object location within the environment. We also experimented with a peaky object distribution, where each object category was assigned a different table to be spawned exclusively on. We present the results in Table \ref{tab:peak-1} and the observations are similar to those reported in Section \ref{s:experiments}.

\renewcommand{\tabcolsep}{1pt}
\begin{table}[h]
\vspace{0.2in}
\centering
\small
\begin{tabular}{c|ccccccc|}
\cline{2-8}
 &
  \multicolumn{7}{c|}{\textbf{Kitchen Environment 1 (Peaky distributions)}} \\ \cline{2-8} 
 &
  \multicolumn{1}{c|}{} &
  \multicolumn{2}{c|}{\textbf{Gen-Lin}} &
  \multicolumn{2}{c|}{\textbf{Neural}} &
  \multicolumn{2}{c|}{\cellcolor[HTML]{C0C0C0}\textbf{GT-Scores}} \\ \cline{3-8} 
 &
  \multicolumn{1}{c|}{\multirow{-2}{*}{\textbf{TSP}}} &
  \multicolumn{1}{c|}{\textbf{Greedy}} &
  \multicolumn{1}{c|}{\textbf{CPSAT}} &
  \multicolumn{1}{c|}{\textbf{Greedy}} &
  \multicolumn{1}{c|}{\textbf{CP-SAT}} &
  \multicolumn{1}{c|}{\cellcolor[HTML]{C0C0C0}\textbf{Greedy}} &
  \cellcolor[HTML]{C0C0C0}\textbf{CP-SAT} \\ \hline
\multicolumn{1}{|c|}{\textbf{Train SPL}} &
  \multicolumn{1}{c|}{-} &
  \multicolumn{1}{c|}{$0.39$} &
  \multicolumn{1}{c|}{$0.37$} &
  \multicolumn{1}{c|}{$0.35$} &
  \multicolumn{1}{c|}{$0.28$} &
  \multicolumn{1}{c|}{\cellcolor[HTML]{C0C0C0}-} &
  \cellcolor[HTML]{C0C0C0}- \\ \hline
\multicolumn{1}{|c|}{\textbf{Eval Succ Rate}} &
  \multicolumn{1}{c|}{$0.88$} &
  \multicolumn{1}{c|}{$0.9$} &
  \multicolumn{1}{c|}{$0.86$} &
  \multicolumn{1}{c|}{$0.85$} &
  \multicolumn{1}{c|}{$0.88$} &
  \multicolumn{1}{c|}{\cellcolor[HTML]{C0C0C0}$0.88$} &
  \cellcolor[HTML]{C0C0C0}$0.91$ \\ \hline
\multicolumn{1}{|c|}{\textbf{Eval SPL}} &
  \multicolumn{1}{c|}{$0.40$} &
  \multicolumn{1}{c|}{$0.46$} &
  \multicolumn{1}{c|}{$0.55$} &
  \multicolumn{1}{c|}{$0.41$} &
  \multicolumn{1}{c|}{$0.45$} &
  \multicolumn{1}{c|}{\cellcolor[HTML]{C0C0C0}$0.56$} &
  \cellcolor[HTML]{C0C0C0}$0.65$ \\ \hline
\end{tabular}
\caption{Experimental comparison of performance of our agents on a peaky object distribution in Kitchen Environment 1 against the metrics mentioned in Section \ref{s:experiments}.}
\label{tab:peak-1}
\vspace{0.0in}
\end{table}
%
%
%
%
%
\begin{figure}[h]
    \centering
    \includegraphics[width=0.6\textwidth]{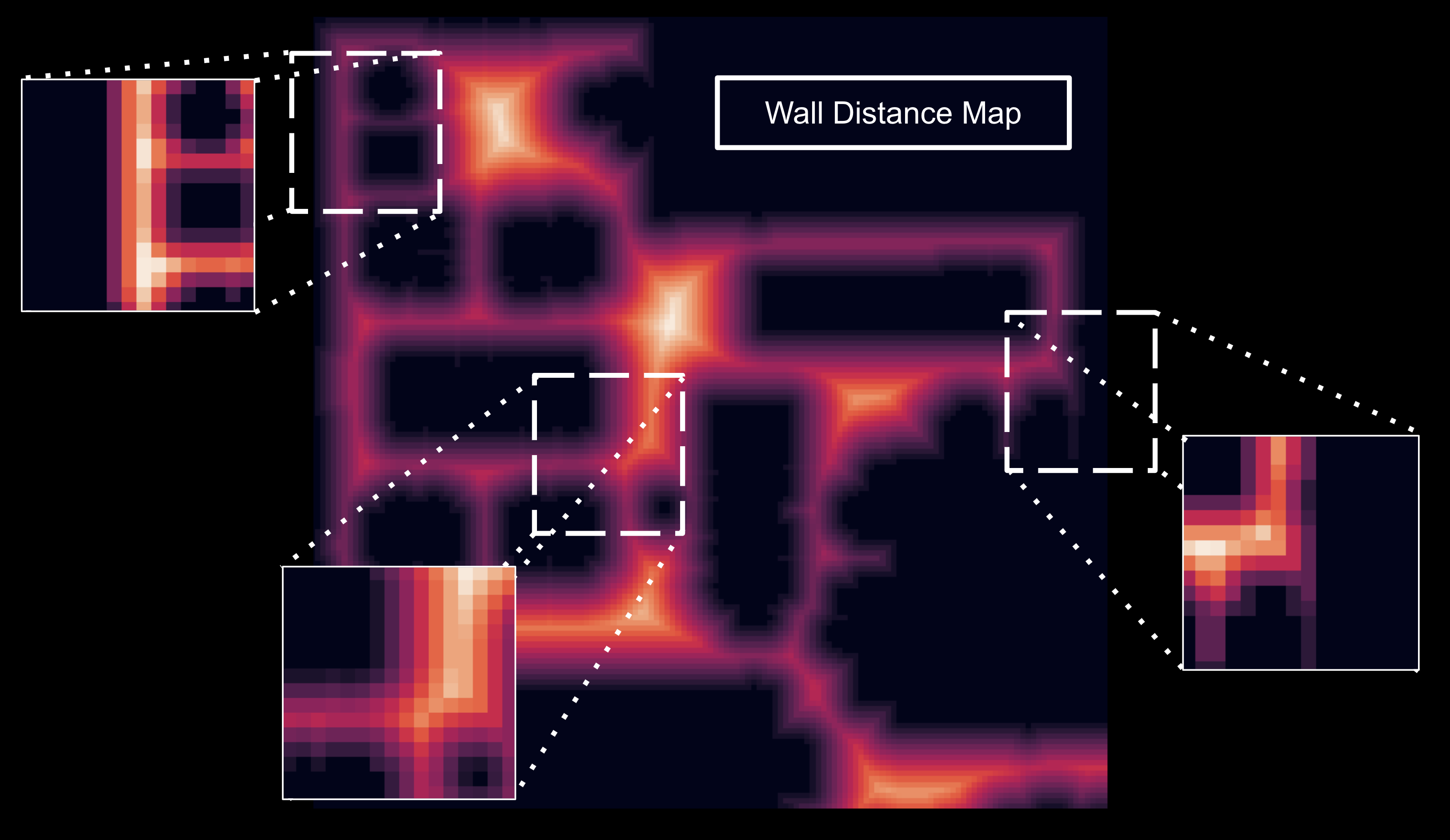}
    \caption{The image shows grid based feature space where a local image patch from various places in the wall distance map are shown. The dotted white boxes indicated from which region the patch features are coming and the patches show how the features look when scaled to a $16 \times 16$ grid.}
    \label{fig:wall_dist_ft}
\end{figure}

\section{Hyper-parameters} \label{hyperparameters}
Tables \ref{tab:hyp-non-peak-1}, \ref{tab:hyp-non-peak-2} and \ref{tab:hyp-peak-1} contain the hyperparameters for each of the algorithms tested in our experiments. Table \ref{tab:hyp-range} gives the values searched for each hyperparameter.

\begin{table}[t!]
\centering
\small
\caption{Hyperparameters for experiments with Non-Peaky object distributions in Kitchen Environment 1.}
\begin{tabular}{|l|cc|cc|}
\hline
\multicolumn{1}{|c|}{}           & \multicolumn{2}{c|}{\textbf{Gen-Lin}} & \multicolumn{2}{c|}{\textbf{Neural}} \\ \hline
\multicolumn{1}{|c|}{\textbf{Hyperparameter}} & \multicolumn{1}{c|}{Greedy}     & CP-SAT     & \multicolumn{1}{c|}{Greedy}     & CP-SAT     \\ \hline
Learning Rate ($\eta$)           & \multicolumn{1}{c|}{$0.44$}  & $100$  & \multicolumn{1}{c|}{$0.01$}   & $0.01$   \\ \hline
Exploration parameter ($\alpha$) & \multicolumn{1}{c|}{$0.1$}   & $0.1$  & \multicolumn{1}{c|}{$0.1$}    & $3.44$   \\ \hline
Number of vantage points ($k$)   & \multicolumn{1}{c|}{$25$}    & $25$   & \multicolumn{1}{c|}{$25$}     & $25$     \\ \hline
Alpha planner ($\alpha_p$)       & \multicolumn{1}{c|}{$0.48$}  & -      & \multicolumn{1}{c|}{$0.43$}   & -        \\ \hline
Map Resolution                   & \multicolumn{1}{c|}{$75$}    & $75$   & \multicolumn{1}{c|}{$75$}     & $75$     \\ \hline
Positional Embedding Size        & \multicolumn{1}{c|}{$50$}    & $15$   & \multicolumn{1}{c|}{$50$}     & $10$     \\ \hline
Feature Vector Normalization                  & \multicolumn{1}{c|}{$l^2$-norm} & $l^2$-norm & \multicolumn{1}{c|}{$l^2$-norm} & $l^2$-norm \\ \hline
Sigmoid Scale ($s$)                    & \multicolumn{1}{c|}{$1.0$}   & $1.0$  & \multicolumn{1}{c|}{$20.0$}   & $19.8$   \\ \hline
\end{tabular}
\label{tab:hyp-non-peak-1}
\end{table}

\begin{table}[t!]
\centering
\small
\caption{Hyperparameters for experiments with Non-Peaky object distributions in Kitchen Environment 2.}
\begin{tabular}{|l|cc|cc|}
\hline
\multicolumn{1}{|c|}{}                        & \multicolumn{2}{c|}{\textbf{Gen-Lin}}    & \multicolumn{2}{c|}{\textbf{Neural}} \\ \hline
\multicolumn{1}{|c|}{\textbf{Hyperparameter}} & \multicolumn{1}{c|}{Greedy}   & CP-SAT   & \multicolumn{1}{c|}{Greedy}   & CP-SAT   \\ \hline
Learning Rate ($\eta$)            & \multicolumn{1}{c|}{$16.62$} & $10.44$ & \multicolumn{1}{c|}{$0.01$}  & $0.01$  \\ \hline
Exploration parameter ($\alpha$)    & \multicolumn{1}{c|}{$10$}    & $0.1$   & \multicolumn{1}{c|}{$0.1$}   & $2.04$  \\ \hline
Number of vantage points ($k$) & \multicolumn{1}{c|}{$50$}    & $50$    & \multicolumn{1}{c|}{$50$}    & $50$    \\ \hline
Alpha planner ($\alpha_p$)         & \multicolumn{1}{c|}{$0.49$}  & -       & \multicolumn{1}{c|}{$0.38$}  & -       \\ \hline
Map Resolution               & \multicolumn{1}{c|}{$37$}    & $37$    & \multicolumn{1}{c|}{$75$}    & $75$    \\ \hline
Positional Embedding Size    & \multicolumn{1}{c|}{$20$}    & $50$    & \multicolumn{1}{c|}{$50$}    & $15$    \\ \hline
Feature Vector Normalization                  & \multicolumn{1}{c|}{mean-var} & mean-var & \multicolumn{1}{c|}{mean-var} & mean-var \\ \hline
Sigmoid Scale ($s$)               & \multicolumn{1}{c|}{-}       & -       & \multicolumn{1}{c|}{$10.00$} & $17.56$ \\ \hline
\end{tabular}
\label{tab:hyp-non-peak-2}
\end{table}

\begin{table}[t!]
\centering
\small
\caption{Hyperparameters for experiments with Peaky object distributions in Kitchen Environment 1.}
\begin{tabular}{|l|cc|cc|}
\hline
\multicolumn{1}{|c|}{}                        & \multicolumn{2}{c|}{\textbf{Gen-Lin}}        & \multicolumn{2}{c|}{\textbf{Neural}}     \\ \hline
\multicolumn{1}{|c|}{\textbf{Hyperparameter}} & \multicolumn{1}{c|}{Greedy}     & CP-SAT     & \multicolumn{1}{c|}{Greedy}     & CP-SAT     \\ \hline
Learning Rate ($\eta$)         & \multicolumn{1}{c|}{$3.98$} & $75.41$ & \multicolumn{1}{c|}{$0.01$}  & $0.01$  \\ \hline
Exploration parameter ($\alpha$)              & \multicolumn{1}{c|}{$7.15$}     & $2.59$     & \multicolumn{1}{c|}{$0.10$}     & $1.40$     \\ \hline
Number of vantage points ($k$) & \multicolumn{1}{c|}{$25$}   & $25$    & \multicolumn{1}{c|}{$25$}    & $25$    \\ \hline
Alpha planner ($\alpha_p$)     & \multicolumn{1}{c|}{$0.59$} & -       & \multicolumn{1}{c|}{$0.57$}  & -       \\ \hline
Map Resolution                 & \multicolumn{1}{c|}{$75$}   & $75$    & \multicolumn{1}{c|}{$75$}    & $75$    \\ \hline
Positional Embedding Size      & \multicolumn{1}{c|}{$10$}   & $20$    & \multicolumn{1}{c|}{$10$}    & $10$    \\ \hline
Feature Vector Normalization                  & \multicolumn{1}{c|}{$l^2$-norm} & $l^2$-norm & \multicolumn{1}{c|}{$l^2$-norm} & $l^2$-norm \\ \hline
Sigmoid Scale                  & \multicolumn{1}{c|}{-}      & -       & \multicolumn{1}{c|}{$20.00$} & $10.67$ \\ \hline
\end{tabular}
\label{tab:hyp-peak-1}
\end{table}

\begin{table}[]
\centering
\small
\caption{Hyperparameter for experiments in the real world.}
\begin{tabular}{|l|c|c|}
\hline
\multicolumn{1}{|c|}{\textbf{Hyperparameter}} & Greedy    & CP-SAT    \\ \hline
Learning Rate ($\eta$)                        & $0.01$    & $0.01$    \\ \hline
Exploration parameter ($\alpha$)              & $0.11$    & $0.1$     \\ \hline
Number of vantage points ($k$)                & $25$      & $25$      \\ \hline
Alpha planner ($\alpha_p$)                    & $0.49$    & -         \\ \hline
Map Resolution                                & $75$      & $75$      \\ \hline
Positional Embedding Size                     & $20$      & $30$      \\ \hline
Feature Vector Normalization                  & $l^2\text{-norm}$ & $l^2\text{-norm}$ \\ \hline
Sigmoid Scale                                 & $18.38$   & $15.78$   \\ \hline
\end{tabular}
\label{tab:hyp-real}
\end{table}

\begin{table}[]
\centering
\small
\caption{Hyperparameter search ranges and scales.}
\label{tab:hyp-range}
\begin{tabular}{|l|c|c|}
\hline
\multicolumn{1}{|c|}{\textbf{Hyperparameter}} & \textbf{Values Searched}                  & \textbf{Search Scale} \\ \hline
Learning Rate ($\eta$) (Gen-Lin model only) & $[0.01, 100.0]$             & Log    \\ \hline
Exploration parameter ($\alpha$)            & $[0.1, 10.0]$               & Linear \\ \hline
Number of vantage points ($k$)              & $\{25, 50\}$                & --     \\ \hline
Alpha planner ($\alpha_p$) (Greedy only)    & $[0.1, 0.9]$                & Linear \\ \hline
Map Resolution                              & $\{37, 75, 150\}$           & --     \\ \hline
Positional Embedding Size                   & $\{5, 10, 15, 20, 30, 50\}$ & --     \\ \hline
Feature Vector Normalization                  & $\{l^2\text{-norm}, \text{mean-var}\}$ & --                    \\ \hline
Sigmoid Scale (for Neural model only)       & $[10, 20]$                  & Linear \\ \hline
\end{tabular}
\end{table}


\begin{figure}
    \centering
    \includegraphics[width=0.6\linewidth]{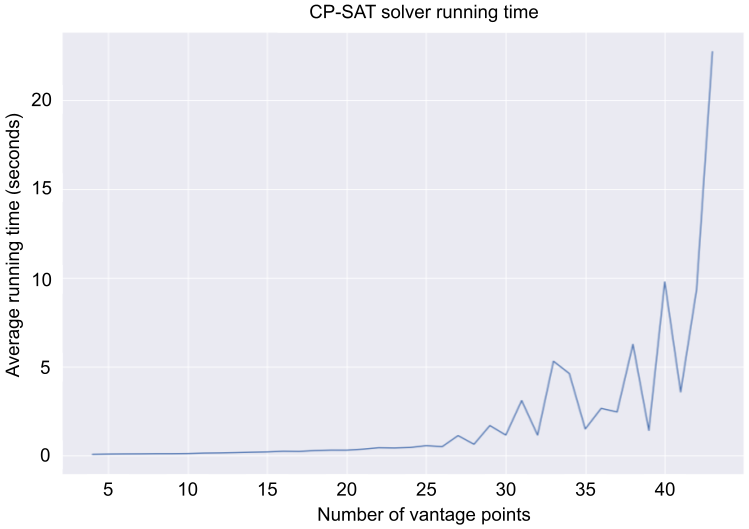}
    \caption{Running time of CP-SAT solver on Intel Xeon 8-core CPU for different numbers of vantage points in Kitchen Environment 2 environment.}
    \label{fig:cpsat-running-time}
\end{figure}
\end{document}